\documentclass[10pt,twocolumn,letterpaper]{article}

\usepackage{cvpr}              

\usepackage[dvipsnames]{xcolor}
\newcommand{\red}[1]{{\color{red}#1}}

\definecolor{cvprblue}{rgb}{0.21,0.49,0.74}
\usepackage[pagebackref,breaklinks,colorlinks,citecolor=cvprblue]{hyperref}
\usepackage{algorithm}
\usepackage{algorithmic}
\usepackage{graphicx}
\usepackage{amsmath}
\usepackage{amssymb}
\usepackage{booktabs}
\usepackage{xcolor,colortbl}
\usepackage{color, colortbl}
\definecolor{tabhighlight}{HTML}{e5e5e5}
\definecolor{citecolor}{HTML}{0071bc}
\newcommand{\tableCellHeight}{1}
\newcommand{\tabstyle}[1]{
  \setlength{\tabcolsep}{#1}
  \renewcommand{\arraystretch}{\tableCellHeight}
  \centering
  \small
}
\usepackage{subcaption}
\usepackage{sidecap}
\newcommand{\tablestyle}[2]{\setlength{\tabcolsep}{#1}\renewcommand{\arraystretch}{#2}\centering\footnotesize}
\usepackage{color}
\def\blu#1{\textbf{\color{blue} #1}} 
\def\red#1{\textbf{\color{red}  #1}} 
\usepackage{newfloat}
\usepackage{listings}
\usepackage[capitalize]{cleveref}
\crefname{section}{Sec.}{Secs.}
\Crefname{section}{Section}{Sections}
\Crefname{table}{Table}{Tables}
\crefname{table}{Tab.}{Tabs.}
\DeclareCaptionStyle{ruled}{labelfont=normalfont,labelsep=colon,strut=off} 
\floatstyle{ruled}
\newfloat{listing}{tb}{lst}{}
\floatname{listing}{Listing}
\def\paperID{5284} 
\def\confName{CVPR}
\def\confYear{2024}

\title{Domain Prompt Learning with Quaternion Networks}

\author{
    Qinglong Cao\textsuperscript{\rm 1,2},
    Zhengqin Xu\textsuperscript{\rm 1},
    Yuntian Chen\textsuperscript{\rm 2}\thanks{Corresponding Author},
    Chao Ma\textsuperscript{\rm 1}\footnotemark[1],
    Xiaokang Yang\textsuperscript{\rm 1}
   \\ $^1$ MoE Key Lab of Artificial Intelligence, AI Institute, Shanghai Jiao Tong University, Shanghai, China\\
$^2$ Ningbo Institute of Digital Twin, Eastern Institute of Technology, Ningbo, China\\
{\tt\small \{caoql2022, fate311\}@sjtu.edu.cn, ychen@eitech.edu.cn, \{chaoma, xkyang\}@sjtu.edu.cn}
}

\begin{document}
\maketitle
\begin{abstract}

Prompt learning has emerged as an effective and data-efficient technique in large Vision-Language Models (VLMs). However, when adapting VLMs to specialized domains such as remote sensing and medical imaging, domain prompt learning remains underexplored. While large-scale domain-specific foundation models can help tackle this challenge, their concentration on a single vision level makes it challenging to prompt both vision and language modalities. To overcome this, we propose to leverage domain-specific knowledge from domain-specific foundation models to transfer the robust recognition ability of VLMs from generalized to specialized domains, using quaternion networks. Specifically, the proposed method involves using domain-specific vision features from domain-specific foundation models to guide the transformation of generalized contextual embeddings from the language branch into a specialized space within the quaternion networks. Moreover, we present a hierarchical approach that generates vision prompt features by analyzing intermodal relationships between hierarchical language prompt features and domain-specific vision features. In this way, quaternion networks can effectively mine the intermodal relationships in the specific domain, facilitating domain-specific vision-language contrastive learning. Extensive experiments on domain-specific datasets show that our proposed method achieves new state-of-the-art results in prompt learning.
\end{abstract}    
\section{Introduction}
\label{sec:intro}

%
Supervised learning using large-scale training samples has shown remarkable success in various visual understanding tasks~\cite{simonyan2014very, he2016deep, long2015fully, redmon2016you}. However, conventional supervised learning often requires a significant amount of data. To address this issue, large-scale Vision-Language Models (VLMs)~\cite{radford2021learning, su2019vl, jia2021scaling} have recently emerged, making prompt learning a more popular and efficient paradigm for various vision tasks. Prompt learning~\cite{zhou2022learning, rao2022denseclip, luddecke2022image, wang2022learning} focuses on adapting VLMs effectively for downstream tasks with limited training data.

\begin{figure}[t]
	\begin{center}
	\end{center}
		
 \includegraphics[width=.90\linewidth]{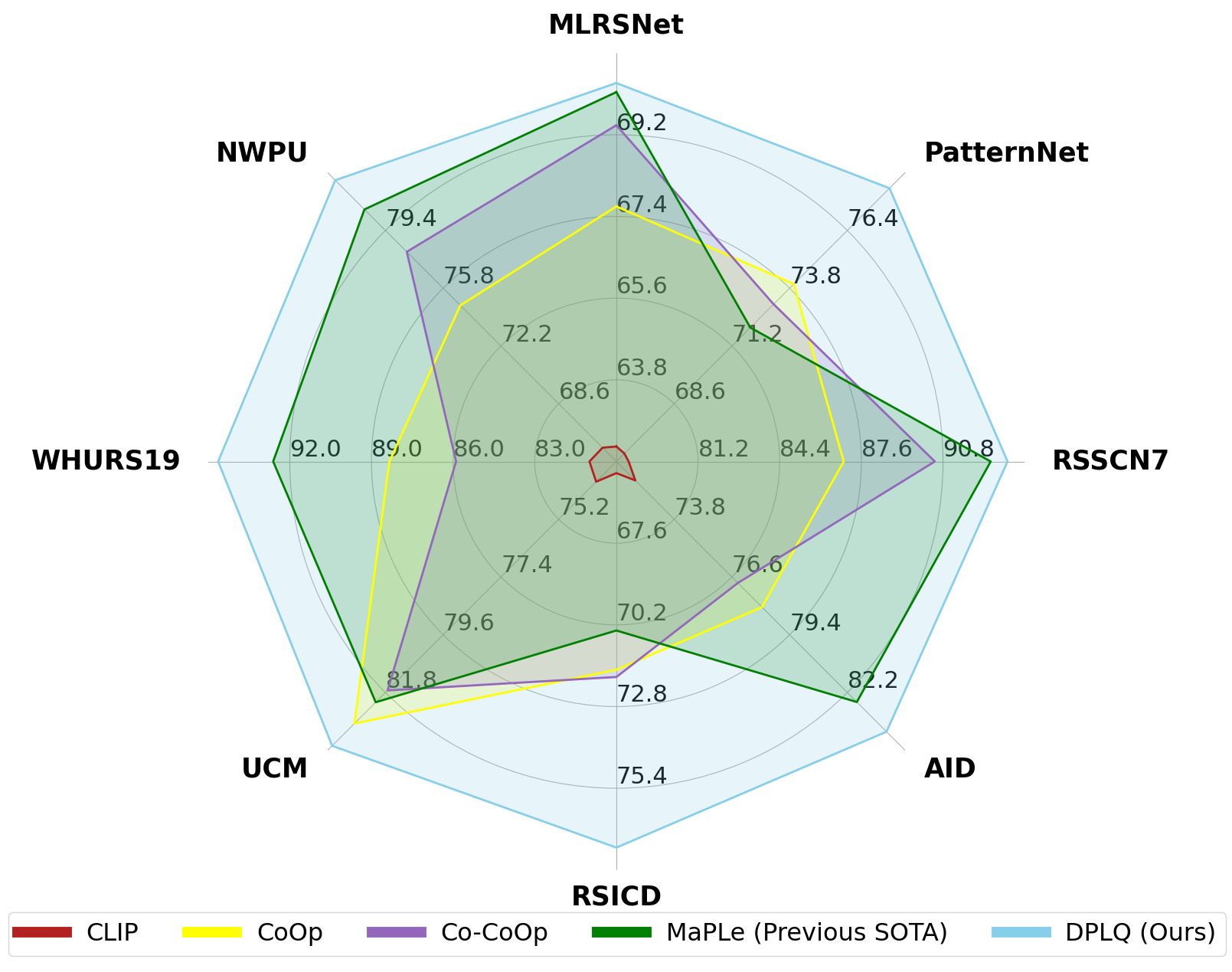}
  \vspace{-3mm}
	\caption{A comparison of the novel class generalization performance of our method against existing methods. Our method achieves state-of-the-art performance on various remote sensing image datasets in terms of harmonic mean. }
 \label{fig1}
 \vspace{-8mm}
\end{figure}


Current VLMs have achieved remarkable progress in contrastive learning with extensive image-text pairs. Among these models, Contrastive Language-Image Pretraining (CLIP)~\cite{radford2021learning} is widely recognized for its exceptional zero-shot generalization capabilities. However, the reliance on fixed prompts challenges the adaptability of CLIP to downstream tasks. CoOp~\cite{zhou2022learning} introduces context optimization by using learnable context vectors as language prompts to address this limitation, which enhances the adaptability of CLIP for visual recognition tasks. However, challenges related to class shifting remain. Thus, CoCoOp~\cite{zhou2022conditional} further introduces input-conditional prompts for the language branch to refine the alignment of both vision and language branches. MaPLe~\cite{khattak2023maple} goes further by introducing a multi-modal prompt learning approach. MaPLe fine-tunes vision and language branches concurrently, leading to more harmonious alignment and improved performance.

It is worth noting that these prompt learning techniques primarily focus on natural images, and these models without domain-specific knowledge tend to misinterpret specific domain images in natural image patterns. Consequently, the existing adaptation of VLMs to specific domains such as remote sensing and medical imaging would be compromised and further result in a significant performance gap. Large-scale domain-specific foundation model~\cite{wang2022advancing,sun2022ringmo,ma2023segment} could be leveraged to address this challenge. However, existing domain-specific foundation models are only pre-trained at the vision level, lacking inherent support for prompting vision-language pairs in a contrastive learning framework.

Our motivation comes from quaternion networks~\cite{parcollet2018quaternion}, known for their effective orthogonal relation modeling and powerful exploration of inter- and intra-correlations within the quaternion hidden space. We leverage the domain-specific knowledge from domain-specific foundation models to transfer the strong recognition ability of VLMs from generalized to specialized domains with quaternion networks. 
Our method propagates the domain-specific vision features from domain-specific foundation models and the generalized contextual embeddings from the language branch into the quaternion hidden space. Within this space, the quaternion network mines the domain-specific cross-modal knowledge for the language branch and projects the generalized contextual embeddings into the specialized space. To address potential overfitting concerns~\cite{zhou2022conditional} in the prompt learning process, we introduce random noise into the quaternion hidden space, enhancing the robustness of the learning process.

Pre-trained VLMs establish a strong and consistent vision-language matching relationship. Thus, we can easily forward the domain-specific information from the specialized language branch into the vision branch. We utilize the learnable language prompt feature as input and the orthogonal domain-specific vision features as guidance to mine the intermodal relations in each vision-language layer, which hierarchically provides vision prompt features. As shown in Figure~\ref{fig1}, 
both the vision and language branches become domain-specific, and the domain-specific contrastive learning minimizes the domain gap, leading to uncompromised domain-specific recognition performance.

In summary, we make the following contributions:
\begin{itemize}

\item To our knowledge, we first introduce the quaternion concept into prompt learning for specific domains. It successfully transfers the strong recognition ability of VLMs from the generalized domain to specialized fields such as remote sensing and medical images.

\item We forward domain-specific info of pre-trained VLMs hierarchically to the vision branch via quaternion hidden space, enabling better recognition performance.
	
\item We extensively validate our method on large-scale domain-specific datasets, such as remote sensing imagery and medical images. Our method achieves state-of-the-art performance.	
\end{itemize}

\section{Related Work}
\label{sec:formatting}
\noindent \textbf{Vision Language Models.}
Vision Language Models (VLMs) aim to establish a stable connection between visual content and textual descriptions by creating a unified embedding space encompassing visual and linguistic modalities. These models can generally be classified into three main groups: models with contrastive objectives~\cite{radford2021learning, mu2022slip, li2021supervision}, generative objectives~\cite{bao2021beit,ko2022large}, and alignment objectives~\cite{singh2022flava,dou2022coarse}. Among these models, CLIP~\cite{radford2021learning} pioneered a symmetrical image-language contrastive loss, enabling robust zero-shot prediction capabilities for various downstream tasks. However, its effectiveness heavily relies on the availability of large and expensive image-text paired datasets. To mitigate this dependency, ALIGN~\cite{jia2021scaling} leverages large-scale noisy image-text pairs, achieving comparable performance through a noise-robust contrastive learning approach.
Subsequently, various researchers have explored more efficient VL model pre-training with fewer image-text pairs. For example, OTTER~\cite{wu2021data} uses optimal transport distillation to establish a soft image-text correspondence, enabling efficient zero-shot recognition with significantly reduced training data. DeCLIP~\cite{li2021supervision} harnesses self-supervision, multi-view supervision, and nearest-neighbor supervision to extract valuable information from limited data. ZeroVL~\cite{cui2022contrastive} successfully employs debiased data sampling and coin flipping mixup for efficient contrastive learning.
In contrast to models primarily designed for recognition tasks, there is a separate category of VL models~\cite{yu2022coca,huang2023nlip,chen2022pali} oriented toward captioning tasks. For instance, COCA~\cite{yu2022coca} employs an encoder-decoder architecture to align embeddings between images and their corresponding captions, yielding impressive results. Through utilizing these advanced VL models, many traditional vision tasks like object detection~\cite{du2022learning},  semantic segmentation~\cite{xu2022simple}, and caption~\cite{tang2021clip4caption}, have achieved great progress. Yet, these works are still limited to natural images. In this paper, we creatively transfer the VLMs into the specialized domains with quaternion networks. 


\vspace{2mm}\noindent \textbf{Prompt Learning.} 
VLMs offer adaptability to downstream tasks through full fine-tuning or linear probing methods. However, full fine-tuning can be computationally intensive and put the established cross-modal representations at risk. On the other hand, linear probing has limitations, particularly concerning models with zero-shot capabilities like CLIP. 
In the field of natural language processing~\cite{liu2023pre}, prompt learning techniques~\cite{zhou2022learning,zhou2022conditional,khattak2023maple} have been proposed, which are efficient for VL models. For instance, CoOp~\cite{zhou2022learning} proposes an effective prompt-learning approach where language prompts are modeled with learnable embeddings while preserving the pre-trained parameters. However, the learned context in CoOp might lack generalizability to unseen classes within the same dataset. To overcome this challenge, CoCoOp~\cite{zhou2022conditional} introduces input-conditional tokens for each image, which act as dynamic prompts to mitigate class shift issues and enhance generalization performance. To prevent prompt tuning from causing a loss of general knowledge, ProGrad~\cite{zhu2023prompt}  selectively updates prompts based on their gradients aligning with the pre-defined prompt predictions. 
Despite the promise, existing prompt learning methods mostly focus on the language branch, with limited consideration for the image branch and its depth. To fill this gap, MaPLe~\cite{khattak2023maple} proposes an effective prompting approach where both vision and language branches are aligned simultaneously. However, these prompt learning methods still focus on natural images and few consider the domain-specific adpation problem. To address this issue, we proposed to leverage the domain-specific knowledge from the domain-specific foundation model to achieve domain prompt learning with quaternion networks.
 \begin{figure*}[ht]
	\begin{center}
		\includegraphics[width=.92\linewidth]{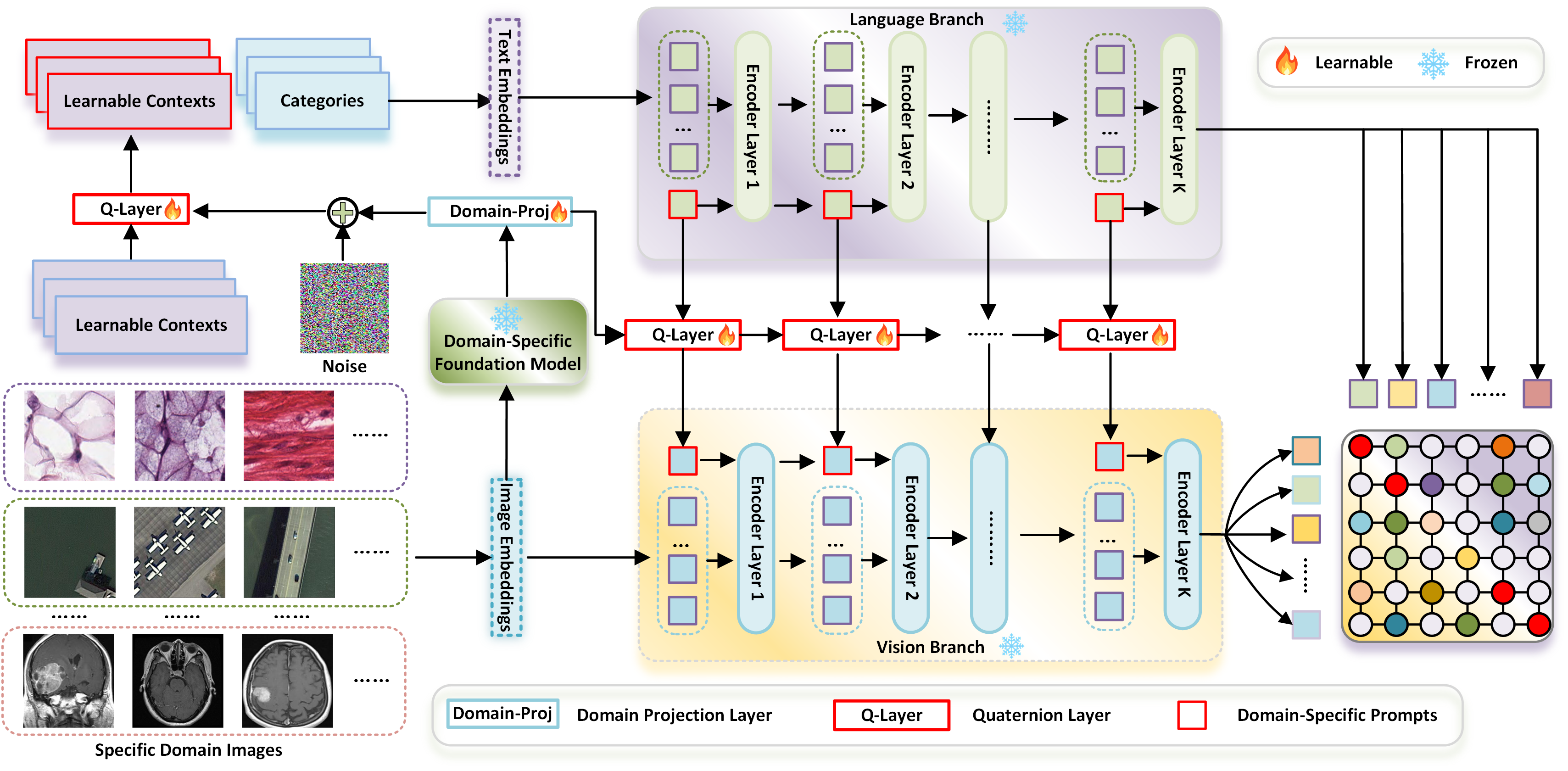}
	\end{center}
        \vspace{-7mm}
	\caption{Overview of our proposed Domain Prompt Learning.  We use the large-scale domain-specific foundation model as guidance, and exploit quaternion networks to mine the intermodal relationships between domain-specific vision features from the domain-specific foundation model and contextual embeddings from the language branch. Based on the stable vision-language matching relationships in pre-trained VLMs, the domain-specific information is hierarchically forwarded from the language branch to the vision branch. }
	\label{fig2}
\end{figure*}

\section{Preliminaries}
\textbf{Quaternion Networks.} 
In four-dimensional space, a quaternion $Q$ extends a complex number and can be expressed as follows:
\begin{equation}
 Q= r1+ x\textbf{i} + y\textbf{j} + z\textbf{k},
\end{equation}
where r, x, y, and z are real numbers, and 1, $\textbf{i},\textbf{j}$, and $\textbf{k}$ are the quaternion unit basis. The real part of $Q$ is denoted by r, while $x\textbf{i} + y\textbf{j} + z\textbf{k}$ is the imaginary or vector part. Quaternions are useful for describing spatial rotations and other applications because they contain embedded information that can be represented by a matrix of real numbers:
\begin{equation}
{Q} = \left[ {\begin{array}{*{20}{c}}
r&{ - x}&{ - y}&{ - z}\\
x&r&{ - z}&y\\
y&z&r&{ - x}\\
z&{ - y}&x&r
\end{array}} \right].
\end{equation}

A quaternion neural network can be defined as:
\begin{equation}
{Q_{out}} = \alpha ( W \otimes Q),
\end{equation}
where $W$ represents the learnable parameters of the quaternion neural networks, $\otimes$ denotes the Hadamard product, and $\alpha$ is the activation function defined as:
\begin{equation}
\alpha (Q) = f(r)1 + f(x)\textbf{i} + f(y)\textbf{j} + f(z)\textbf{k},
\end{equation}
where $f$ is any standard activation function. 
We suggest using quaternion networks to discover orthogonal intermodal relationships between domain-specific vision features from the domain-specific  foundation models and contextual embeddings from the language branch. This approach is inspired by the unique feature processing pattern of the quaternion networks.

\vspace{2mm}\noindent\textbf{Prompt Learning.} 
%
%
The CLIP model consists of a visual encoder and a text encoder that generate image embeddings and corresponding text embeddings, respectively. The CLIP model is based on the vision transformer and follows the same setting as previous methods~\cite{khattak2023maple, zhou2022conditional}. During training, the CLIP model maximizes the cosine similarity between the image and its matched text and minimizes the cosine similarity between the image and its unmatched text. This allows the CLIP model to perform zero-shot classification. 
To perform zero-shot classification, the text embedding $\omega_i$ is generated from a hand-crafted prompt, such as ``a photo of $category$", where $category$ is the $i$-th class name. If there are $C$ categories and the visual embedding of the image is $x$, the probability of the image belonging to the $i$-th class name is given by:
\begin{equation}
\label{eq3}
p(y_i|x) = \frac{{\exp (sim(x,\omega_i )/\tau )}}{{\sum\nolimits_{i = 1}^C {\exp (sim(x,\omega_i )} /\tau )}},
\end{equation}
where $sim()$ is the cosine similarity function and $\tau$ is the temperature hyperparameter.

\section{Proposed Method}
\subsection{Overview}
The aim of this work is to efficiently prompt the VLMs from the generalized domain to specific domains such as remote sensing and medical images. To achieve this, we introduce quaternion networks, which enable the incorporation of domain-specific knowledge from large-scale foundation models into VLMs. As shown in Figure~\ref{fig2},
by utilizing the quaternion network, we are able to identify cross-modal relationships between domain-specific vision features from the domain-specific foundation model and generalized contextual embeddings from the language branch. This information is then used to project the generalized contextual embeddings into the specialized space. Additionally, well-matched vision-language relationships in pre-trained VLMs are utilized to propagate domain-specific information from the specialized language branch into the vision branch. 
Consequently, the proposed domain-specific prompt learning significantly enhances recognition performance.

\subsection{Domain-Specific Foundation Model}

The development of large-scale domain-specific foundation models~\cite{sun2022ringmo,wang2022advancing, ma2023segment} has significantly improved the quality of representations for downstream vision tasks, particularly in remote sensing and medical imaging. In line with this advancement, we incorporate the large-scale remote sensing foundation model~\cite{wang2022advancing} and MedSAM~\cite{ma2023segment} to provide essential domain-specific knowledge for remote sensing and medical images, respectively. The foundation model for remote sensing mainly uses architectures like Vision Transformers~\cite{dosovitskiy2020image} and ViTAE~\cite{xu2021vitae} and trains these networks with millions of remote sensing images using a masked autoencoder~\cite{he2022masked} approach. The masked autoencoder framework is designed to reconstruct masked images, emphasizing the recovery of visible portions within an encoder-decoder architecture. Network optimization involves minimizing the loss between the reconstructed regions and the corresponding ground-truth masked regions. In the medical image processing domain, MedSAM~\cite{ma2023segment} leverages a meticulously curated dataset of over one million medical images for pre-training. To effectively prompt the visual and language branches into the specific domain space, we introduce domain-specific foundation models to provide domain-specific knowledge.

\subsection{Prompting Language Branch}

In order to obtain domain-specific vision features $F_d$, we propagate the image patch embeddings into the domain-specific foundation model. To better represent the domain-specific information in the quaternion hidden space, we apply a domain-projection layer $L_d$ consisting of two linear layers on $F_d$, which results in the projected domain-specific vision features $\widehat F_d$:
\begin{equation}
{\widehat F_d} = {L_d}({F_d}),
\end{equation}

To mine the critical orthogonal intermodal relationship, we model the domain-specific vision features $\widehat F_d$ and the learnable context embeddings $T_c$ in two orthogonal axes in the quaternion hidden space:
\begin{equation}
Q_l = T_c + {\widehat F_d}\textbf{i} + 0\textbf{j} + 0\textbf{k},
\end{equation}

To support the quaternion projection, we construct a zero tensor $Z_0$ with the same size of $T_c$. However, as mentioned in CoCoOp~\cite{zhou2022conditional}, prompt learning is prone to the problem of overfitting due to the limited amount of data involved. To address this issue, we add some random Gaussian noise $N_G$ into the quaternion hidden space to implicitly enhance the robustness of the learning process. The noise is scaled by the mean of the domain-specific vision features $\widehat F_d$:
\begin{equation}
    N_G = Mean(\widehat F_d) N_{\theta},
\end{equation}
where $N_{\theta}$ denotes the standard Gaussian noise. Given quaternion layer $Q_t$, the generation of domain-specific context embedding $T_d$ is computed as follows:
\begin{equation}
    T_d = Q_t([{\widehat F_d}+T_c+N_G, Z_0]),
\end{equation}

In this way, the orthogonal intermodal relationship between domain-specific vision features and contextual embeddings is well-mined in the quaternion hidden space, and the generalized contextual embeddings are projected into the specialized space. 
To prompt the visual encoder with the domain-specific information, we leverage domain-specific context embedding $T_d$ and add a set of learnable language prompt features $\left[ {P_l^1, P_l^2,..., P_l^m} \right]$ with the setting depth of $k$ into encoder layers $\left[ {L_l^1, L_l^2,..., L_l^m} \right]$ of the language branch, where $m$ denotes the total layer numbers. 
Given the fixed text embedding $C_t$ for categories, we first concatenate $T_d$ with $C_t$ to acquire the complete domain-specific text embeddings, i.e., $W_1 = \left[T_d, C_t \right]$. The domain-specific information propagation could then be computed as follows:
\begin{equation}
\left[ {{W_i},~~\_~~} \right] = {L_l^i}([{W_{i-1}},P_l^{i-1}])~~~~~i = 1,...,k,
\end{equation}
\begin{equation}
\left[ {{W_j},P_l^j} \right] = {L_l^j}([{W_{j-1}},P_l^{j-1}])~~~~~j = k+1,...,m,
\end{equation}

\subsection{Prompting Vision Branch}

The pre-trained VLMs establish a solid vision-language matching relationship, which allows us to easily transfer domain-specific information from the language branch to the vision branch. To achieve this, we set a group of learnable vision prompt features, denoted by $\left[ {P_v^1, P_v^2,..., P_v^m} \right]$, with a depth of $k$, which correspond to the encoder layers $\left[ {L_v^1, L_v^2,..., L_v^m} \right]$ of the vision branch. Next, we introduce a group of quaternion layers $\left[ {Q_v^1, Q_v^2,..., Q_v^m} \right]$ with a depth of $k$, which are responsible for providing cross-modal domain-specific information.
In a similar way to how the quaternion computation is performed in the language branch, domain-specific vision features $\widehat F_d$ and language prompt features $P_l^i$ are modeled in two orthogonal axes in the quaternion hidden space for the vision branch. We use the following equation to perform this computation:
\begin{equation}
Q_v^i = P_l^i + {\widehat F_d}\textbf{i} + 0\textbf{j} + 0\textbf{k},
\end{equation}

As the vision-language relationship is well-matched, random noise is no longer required. Therefore, we can compute the vision prompt features as follows:
\begin{equation}
    P_v^i = Q_v^i([{\widehat F_d}+P_l^i, Z_0])~~~~~~i = 1,2,...,k,
\end{equation}

By leveraging the well-matched vision-language relationship, we propagate the domain-specific knowledge into the vision prompt features within the quaternion hidden space.
To propagate domain-specific information for the vision branch, we start with the original image embeddings $E_1$ and class token $c_1$, and compute the propagation as follows:
\begin{equation}
\left[ {{E_i}, c_i, ~~\_~~} \right] = {L_v^i}([{E_{i-1}},{c_{i-1}}, P_v^{i-1}]) ~i = 1,...,k,
\end{equation}

We then compute the following equation to propagate the domain-specific information further:
\begin{equation}
\begin{split}
\left[ {{E_j},c_i, P_v^j} \right] &= {L_v^j}([{E_{j-1}}, {c_{j-1}}, P_v^{j-1}]) \\
& j = k+1,...,m,
\end{split}
\end{equation}

Finally, given vision embeddings $E_m$ of the last vision layer and language embeddings $\left[W^1_m,W^2_m,..., W^C_m,\right]$ for $C$ categories of the last language layer, we compute the probability of an image belonging to a specific category as follows:  
\begin{equation}
p(y_i|x) = \frac{{\exp (sim({E_m},W_m^i)/\tau )}}{{\sum\nolimits_{j = 1}^C {\exp (sim({E_m},W_m^j)/\tau )} }},
\end{equation}
where $p(y_i|x)$ is the probability of the image belonging to the $i$-th category, $sim$ denotes the similarity function, and $\tau$ is a temperature parameter.

\begin{table*}[t!]
\small
\tablestyle{6pt}{0}
\addtolength{\tabcolsep}{-6pt}
    \tabstyle{1.0pt}
    \setlength{\tabcolsep}{3.2pt}
    \caption{Comparison with SOTA methods in base-to-novel generalization on 8 remote sensing recognition datasets. Our method consistently performs well over the SOTA approaches. We use \red{red} and \blu{blue} to highlight the first and second best scores.
    }
    \vspace{-2mm}
    \scalebox{0.9}{
    \begin{subtable}[t]{.32\textwidth}
    \centering
    \caption{\textbf{Average over 8 datasets}}
    \begin{tabular}{l cc c}
    \toprule
    & Base & Novel & HM \\
    \midrule
    CLIP~\cite{radford2021learning} & 71.19	&71.33	&70.63 \\
    CoOp~\cite{zhou2022learning} & 87.61	&70.84	&78.03 \\
    CoCoOp~\cite{zhou2022conditional} & 91.82	&68.98	&78.43 \\
    MaPLe~\cite{khattak2023maple} & 93.12	&71.71	&80.42 \\
    \midrule
    Ours (ViTAE) &\red{94.28}	&\blu{73.43}	&\blu{82.05}
\\
    \rowcolor{tabhighlight}
       Ours (ViT) & \blu{94.08} &\red{75.06}	&\red{83.50}
\\
    \bottomrule
    \end{tabular}
    \end{subtable}
    }
        \scalebox{0.9}{
    \begin{subtable}[t]{.32\textwidth}
    \centering
    \caption{MLRSNet}
    \begin{tabular}{l cc c}
    \toprule
    & Base & Novel & HM \\
    \midrule
    CLIP~\cite{radford2021learning} & 64.50 & \red{60.30} & 62.33 \\
    CoOp~\cite{zhou2022learning} & 79.37 & 58.90 & 67.62\\
    CoCoOp~\cite{zhou2022conditional} & 83.30 & 59.50 & 69.42 \\
    MaPLe~\cite{khattak2023maple} & 85.23 & \blu{59.60} & \blu{70.15} \\
    \midrule
    Ours (ViTAE) & \red{88.96}	&57.10	&69.56 \\
    \rowcolor{tabhighlight}
     Ours (ViT) &\blu{87.07}	&59.00	&\red{70.34}\\
    \bottomrule
    \end{tabular}
    \end{subtable}
    }
    ~
    \scalebox{0.9}{
    \begin{subtable}[t]{.32\textwidth}
    \centering
    \caption{PatternNet}
    \begin{tabular}{l cc c}
    \toprule
    & Base & Novel & HM \\
    \midrule
    CLIP~\cite{radford2021learning} & 70.60 & 62.60 & 66.36 \\
    CoOp~\cite{zhou2022learning} & 87.30 & \blu{64.20} & 73.99 \\
    CoCoOp~\cite{zhou2022conditional} & 93.70 & 59.90 & 73.08 \\
    MaPLe~\cite{khattak2023maple} & 95.30 & 57.90 & 72.03 \\
    \midrule
    Ours (ViTAE) &\red{97.07}	&62.37	&\blu{75.94}\\
    \rowcolor{tabhighlight}
     Ours (ViT) & \blu{95.80}	&\red{66.20}	&\red{78.30}\\
    \bottomrule
    \end{tabular}
    \end{subtable}
    }
    ~
        \scalebox{0.9}{
    \begin{subtable}[t]{.32\textwidth}
    \centering
    \caption{RSSCN7}
    \begin{tabular}{l cc c}
    \toprule
    & Base & Novel & HM \\
    \midrule
    CLIP~\cite{radford2021learning} & 66.70 & 95.30 & 78.48 \\
    CoOp~\cite{zhou2022learning} & 84.80  &	89.13 & 86.91 \\
    CoCoOp~\cite{zhou2022conditional} & 90.97	&90.00	&90.48 \\
    MaPLe~\cite{khattak2023maple} & \red{91.67}	&93.70	&92.67 \\
    \midrule

    Ours (ViTAE) &\blu{91.53}	&\blu{94.53}	&\blu{93.01}\\
    \rowcolor{tabhighlight}
    Ours (ViT) &91.20	&\red{95.57}	&\red{93.33}\\
    \bottomrule
    \end{tabular}
    \end{subtable}
        }
        \scalebox{0.9}{
    \begin{subtable}[t]{.32\textwidth}
    \centering
    \caption{AID}
    \begin{tabular}{l cc c}
    \toprule
    & Base & Novel & HM \\
    \midrule
    CLIP~\cite{radford2021learning} & 73.50	&70.40 &71.92 \\
    CoOp~\cite{zhou2022learning} & 87.63	&70.37	&78.06\\
    CoCoOp~\cite{zhou2022conditional} &92.63	&65.73	&76.89 \\
    MaPLe~\cite{khattak2023maple} & 92.73	&74.57	&82.66 \\
    \midrule
    Ours (ViTAE) & \blu{94.03}	&\blu{74.97}	&\blu{83.43} \\
    \rowcolor{tabhighlight}
   Ours (ViT) & \red{94.50}	&\red{75.77}	&\red{84.10} \\
    \bottomrule
    \end{tabular}
    \end{subtable}
    }
    ~
        \scalebox{0.9}{
    \begin{subtable}[t]{.32\textwidth}
    \centering
    \caption{RSICD}
    \begin{tabular}{l cc c}
    \toprule
    & Base & Novel & HM \\
    \midrule
    CLIP~\cite{radford2021learning} & 71.50   &60.20	&65.37 \\
    CoOp~\cite{zhou2022learning} & 88.43	&60.20	&71.63 \\
    CoCoOp~\cite{zhou2022conditional} & 92.37	&58.80	&71.86 \\
    MaPLe~\cite{khattak2023maple} & 93.93	&56.27	&70.38 \\
    \midrule
    Ours (ViTAE) &\blu{94.57}	&\red{65.20}	&\blu{77.19}\\
    \rowcolor{tabhighlight}
    Ours (ViT) & \red{95.67}	&\blu{64.83}	&\red{77.29} \\
    \bottomrule
    \end{tabular}
    \end{subtable}
    }
    ~
        \scalebox{0.9}{
    \begin{subtable}[t]{.32\textwidth}
    \centering
    \caption{UCM}
    \begin{tabular}{l cc c}
    \toprule
    & Base & Novel & HM \\
    \midrule
    CLIP~\cite{radford2021learning} & 80.60	&68.00	&73.77 \\
    CoOp~\cite{zhou2022learning} & 93.60	&\red{74.53}	&82.98 \\
    CoCoOp~\cite{zhou2022conditional}& 95.23	&71.57	&81.72 \\
    MaPLe~\cite{khattak2023maple} & \blu{97.70}	&70.90	&82.17 \\
    \midrule
    Ours (ViTAE) &97.10 	&72.10	&\blu{82.75}\\
    \rowcolor{tabhighlight}
    Ours (ViT) &\red{97.90}	&\blu{73.30}	&\red{83.83}\\
    \bottomrule
    \end{tabular}
    \end{subtable}
    }
        \scalebox{0.9}{
    \begin{subtable}[t]{.32\textwidth}
    \centering
    \caption{WHURS19}
    \begin{tabular}{l cc c}
    \toprule
    & Base & Novel & HM \\
    \midrule
    CLIP~\cite{radford2021learning} & 73.10	&\red{90.80}  &80.99 \\
    CoOp~\cite{zhou2022learning} & 95.20	 &82.40	&88.34 \\
    CoCoOp~\cite{zhou2022conditional} & 97.10	&77.00	&85.89 \\
    MaPLe~\cite{khattak2023maple} & 97.70	&88.03	&92.61 \\
    \midrule
    Ours (ViTAE) &\red{99.40}	&\blu{89.90}	&\blu{94.41}\\
    \rowcolor{tabhighlight}
    Ours (ViT)  &\blu{98.80}	&\red{90.80}	&\red{94.63}\\
    \bottomrule
    \end{tabular}
    \end{subtable}
    }
    ~
        \scalebox{0.9}{
    \begin{subtable}[t]{.32\textwidth}
    \centering
    \caption{NWPU}
    \begin{tabular}{l cc c}
    \toprule
    & Base & Novel & HM \\
    \midrule
    CLIP~\cite{radford2021learning} & 69.00	&63.00	&65.87 \\
    CoOp~\cite{zhou2022learning} & 84.53	&66.97	&74.73 \\
    CoCoOp~\cite{zhou2022conditional} & 89.27	&69.37	&78.07 \\
    MaPLe~\cite{khattak2023maple} & 90.70	&\blu{72.70}	&\blu{80.71}\\
    \midrule
    Ours (ViTAE) &\blu{91.60}	&71.23	&80.14\\
    \rowcolor{tabhighlight}
     Ours (ViT) &\red{91.70}	&\red{75.03}	&\red{82.53} \\
    \bottomrule
    \end{tabular}
    \end{subtable}
    }

    \label{table1}
\end{table*}

\section{Experiments}

In order to assess the effectiveness of our proposed quaternion learning approach, we conducted extensive experiments on remote sensing and medical images. Our experiments covered three distinct problem settings: 1) generalization from known to unknown classes within a dataset, 2) transfer between datasets, and 3) generalization across domains. This section provides a detailed description of the datasets, evaluation metrics, and experimental implementations. We also present a thorough performance analysis, as well as ablation experiments to clarify the effectiveness of our proposed designs.

\subsection{Datasets and Evaluation Metrics}


We evaluate the proposed method on 8 remote sensing datasets, namely MLRSNet~\cite{qi2020mlrsnet}, PatternNet~\cite{zhou2018patternnet}, RSSCN7~\cite{zou2015deep}, AID~\cite{xia2017aid}, RSICD~\cite{lu2017exploring}, UCM~\cite{yang2010bag}, WHURS19~\cite{Dai2011WHURS19}, and NWPU~\cite{cheng2017remote}. Additionally, we evaluated the proposed method on 3 medical datasets, including BTMRI~\cite{BTMRI}, CCBTM~\cite{CCBTM}, and CHMNIST~\cite{CHMNIST}. As with previous methods~\cite{khattak2023maple}, we use accuracy and Harmonic Mean (HM) as evaluation metrics:
\begin{equation}
    HM = \frac{{2 \times Acc_{base} \times Acc_{novel}}}{{Acc_{base} + Acc_{novel}}},
\end{equation}
where $Acc_{base}$ denotes the accuracy of base categories, and $Acc_{novel}$ denotes the accuracy of novel categories. We report the experimental results as an average of three independent runs for better statistical reliability. For the base-to-novel generalization scenario, we conduct experiments on all eight remote sensing datasets and three medical datasets. In addition, we evaluat the proposed method on cross-dataset generalization and domain generalization, where MLRSNet is used as the source dataset, and other remote sensing datasets are used as target datasets.

\subsection{Implementation Details}

For a fair comparison, we use similar training settings in MaPLe. All experiments are conducted using a few-shot training strategy with 16 shots, randomly sampled for each class. We use the pre-trained ViT-B/16 CLIP model for prompt tuning. The network is trained with a batch size of 4 and a learning rate of 0.0035 using the SGD optimizer on a single NVIDIA A100 GPU. We use the template ``a photo of $category$" for the word embeddings. To ensure fair and reliable comparisons, we maintain consistent hyperparameters across all datasets, except for the domain-project layer, designed quaternion layer, and learnable contexts, which had their corresponding dimensions specified for each layer. The domain-project layer follows a Linear-ReLU-Linear architecture, while the designed quaternion layer adheres to the standard quaternion layer structure outlined in \cite{parcollet2018quaternion}.

\subsection{Generalization from Base-to-Novel Classes}

The main goal of prompt learning is to facilitate the effective transfer of pre-trained VLMs to downstream tasks. A major challenge is ensuring robust generalization across base and novel classes. To rigorously evaluate our method against the most critical generalization from base-to-novel classes, we carry out experiments on remote sensing and medical images. For remote sensing images, we use the large-scale remote sensing foundation model~\cite{wang2022advancing}, which offers two distinct pre-trained backbones, ViTAE and ViT. To comprehensively assess performance, we conduct experiments using both backbones and compare the results with advanced methods. Detailed experimental comparisons are shown in Table~\ref{table1}. With the ViT backbone, our method achieves SOTA performance, achieving a 3.08$\%$ average HM improvement. Specifically, it improves the base accuracy from 93.12$\%$ to 94.08$\%$ and impressively boosts the novel accuracy by 3.35$\%$. On the RSICD dataset, Our approach significantly increases the base accuracy from 93.93$\%$ to 95.67$\%$ and the novel accuracy from 60.20$\%$ to 64.83$\%$, with an HM improvement of nearly 6$\%$. These findings underscore the benefits of domain-specific information for RSICD. However, on the challenging MLRSNet dataset, our method achieves only a modest 0.19$\%$ HM improvement, mainly due to limited knowledge of the domain-specific foundation model for MLRSNet. Moreover, while the deeper ViTAE performs better under some settings, it does not surpass ViT overall. These results suggest that deeper architectures might not be ideal for extracting domain-specific information, as redundant parameters can potentially hinder performance.

Additionally, we conduct experiments on medical images using a medical domain-specific foundation model~\cite{ma2023segment} to further assess the efficacy of our proposed prompt learning approach. The experimental results and comparisons with advanced approaches are shown in Table~\ref{table2}. Overall, our method achieves the highest accuracy across datasets. On average, our method attains an accuracy of 74.36$\%$ for the base categories and enhances the accuracy of novel categories from 44.40$\%$ to 44.74$\%$. In terms of HM, our approach brings about nearly 4$\%$ improvement. Interestingly, while our method excels in terms of the accuracy of base categories and HM across datasets, it is not the top performer for novel categories. This observation suggests that existing domain-specific foundation models for medical images, although contributing significantly to improving the accuracy of base categories with domain-specific knowledge, may not yet fully harness their potential for enhancing generalization capabilities. This limitation could be a topic for future research exploration.

\begin{table*}[t!]
\small
\tablestyle{6pt}{0}
\addtolength{\tabcolsep}{-6pt}
    \tabstyle{1.0pt}
    \setlength{\tabcolsep}{3.2pt}
    \caption{Comparison between our method and SOTA methods for base-to-novel generalization on medical image classification datasets. Our method performs well over the compared methods. We use \red{red} and \blu{blue} to indicate the first and second best scores. 
    }
    \vspace{-2mm}
    \scalebox{0.95}{
    \begin{subtable}[t]{.25\textwidth}
    \centering
    \caption{\textbf{Average over datasets}}
        \begin{tabular}{l cc c}
    \toprule
    & Base & Novel & HM \\
    \midrule
    CLIP~\cite{radford2021learning} &49.83	&41.83	&45.18 \\
    CoOp~\cite{zhou2022learning} & 51.59	&43.77	&46.81 \\
    CoCoOp~\cite{zhou2022conditional}  & \blu{64.45}	&43.16	&\blu{49.45} \\
    MaPLe~\cite{khattak2023maple}  & 62.39	&\blu{44.40} &49.01 \\
    \midrule
    \rowcolor{tabhighlight}
       Ours  &  \red{74.36}	&\red{44.74}	&\red{53.36}\\
    \bottomrule
    \end{tabular}

    \end{subtable}
    }
        ~
    \begin{subtable}[t]{.20\textwidth}
    \centering
    \caption{All datasets}
\includegraphics[width=0.85\linewidth]{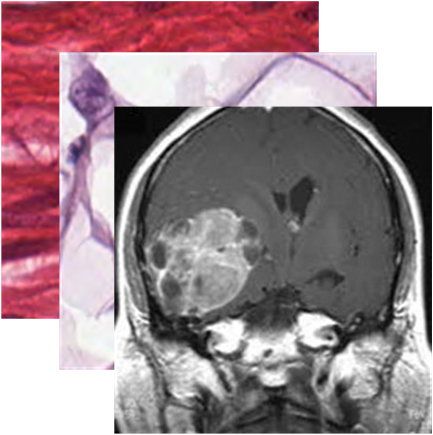}
    \end{subtable}
\scalebox{0.95}{
    \begin{subtable}[t]{.25\textwidth}
    \centering
    \caption{BTMRI}
    \begin{tabular}{l cc c}
    \toprule
    & Base & Novel & HM \\
    \midrule
    CLIP~\cite{radford2021learning}  &50.60	&51.20	&50.89 \\
    CoOp~\cite{zhou2022learning} & 48.93	&53.30	&51.02 \\
    CoCoOp~\cite{zhou2022conditional} & 52.37	&52.80	&52.58 \\
    MaPLe~\cite{khattak2023maple} & \blu{53.67}	&\red{61.60} &\blu{57.36} \\
    \midrule
    \rowcolor{tabhighlight}
       Ours  &\red{60.97}	&\blu{56.30}	&\red{58.54}\\
    \bottomrule
    \end{tabular}
    \end{subtable}
    }
            ~
    \begin{subtable}[t]{.20\textwidth}
    \centering
    \caption{BTMRI dataset}
\includegraphics[width=0.8\linewidth]{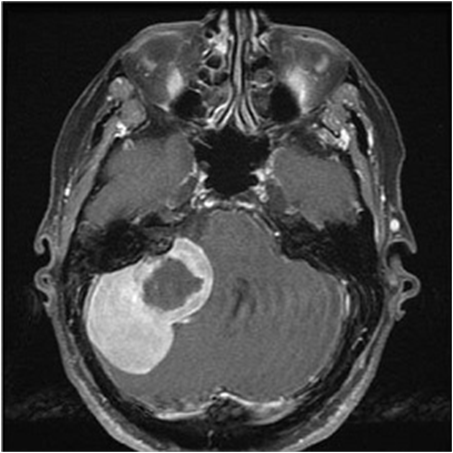}
    \end{subtable}
         ~
        \scalebox{0.95}{
    \begin{subtable}[t]{.25\textwidth}
    \centering
    \caption{CHMNIST}
    \begin{tabular}{l cc c}
    \toprule
    & Base & Novel & HM \\
    \midrule
    CLIP~\cite{radford2021learning} & 31.60 & \red{27.40} & 29.35 \\
    CoOp~\cite{zhou2022learning} & 41.70 & 25.67 & 31.78 \\
    CoCoOp~\cite{zhou2022conditional} & \blu{74.30} & 25.30 & \blu{37.74} \\
    MaPLe~\cite{khattak2023maple} & 74.03 & 25.10 & 37.49 \\
    \midrule
    \rowcolor{tabhighlight}
     Ours  & \red{87.80}	&\blu{26.60}	&\red{40.83}
 \\
    \bottomrule
    \end{tabular}
    \end{subtable}
    }
    ~
    \begin{subtable}[t]{.20\textwidth}
    \centering
    \caption{CHMNIST dataset}
\includegraphics[width=0.8\linewidth]{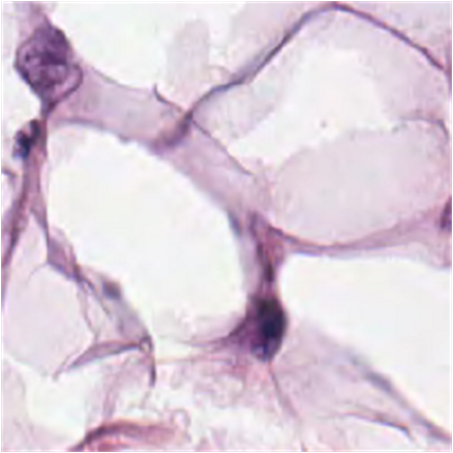}
    \end{subtable}
    ~
        \scalebox{0.95}{
    \begin{subtable}[t]{.25\textwidth}
    \centering
    \caption{CCBTM}
    \begin{tabular}{l cc c}
    \toprule
    & Base & Novel & HM \\
    \midrule
    CLIP~\cite{radford2021learning} & \blu{67.30} & 46.90 & 55.28 \\
    CoOp~\cite{zhou2022learning} & 64.13  &	\red{52.33} & 57.63 \\
    CoCoOp~\cite{zhou2022conditional} & 66.67	&\blu{51.37}	&\blu{58.03} \\
    MaPLe~\cite{khattak2023maple} & 59.47	&46.50	&52.19 \\
    \midrule
    \rowcolor{tabhighlight}
    Ours  & \red{74.30}	&51.33	&\red{60.72}\\
    \bottomrule
    \end{tabular}
    \end{subtable}
        }
        ~
    \begin{subtable}[t]{.20\textwidth}
    \centering
    \caption{CCBTM dataset}
\includegraphics[width=0.8\linewidth]{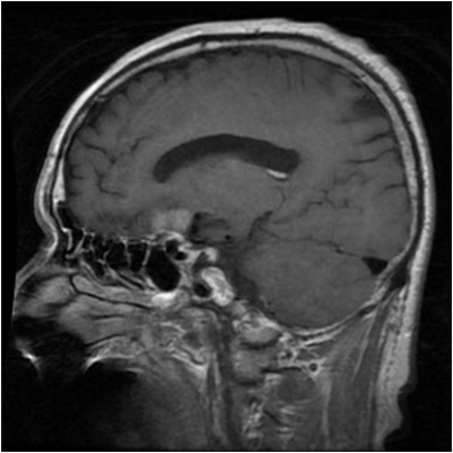}
    \end{subtable}
    \label{table2}
\end{table*}

\begin{table*}[!t]
\small
\centering
    \caption{Comparisons with SOTA methods for cross-dataset generalization with the MLRSNet dataset as the source domain and remaining remote sensing datasets as the target domains. Our method achieves better performance than the compared methods. We use \red{red} and \blu{blue} to highlight the first and second best scores.}
    \vspace{-2mm}
	\scalebox{0.95}{
\begin{tabular}{l c cccccccc}
    \toprule
    & \textbf{\ \ Source\ \ } & \multicolumn{8}{c}{\textbf{Target}} \\
    \cmidrule{2-2} \cmidrule(l){3-10}
   & MLRSNet & PatternNet & RSSCN7 & AID & RSICD & UCM & WHURS19 & NWPU & Average\\
    \midrule
    \ \  CoOp~\cite{zhou2022learning} & 72.53	&66.97	&69.03	&\red{67.30}	&\red{63.50}	&77.57	&\blu{85.47}	&70.43	&71.60\\
    \ \ CoCoOp~\cite{zhou2022conditional} \  & 71.70	&65.67	&68.80	&66.63	&\blu{62.57}	&76.40	&85.33	&70.30	&70.92\\
    \ \ MaPLe~\cite{khattak2023maple} & \blu{76.83}	&\red{68.53}	&\red{71.43}	&65.13	&59.53	&\blu{79.90}	&85.23	&\blu{72.80}	&\blu{72.42}  \\
    \midrule
    \rowcolor{tabhighlight} \ \ Ours & \red{78.73}	&\blu{68.17}	&\blu{70.60}	&\blu{66.70}	&62.27	&\red{79.93}	&\red{91.07}	&\red{73.13}	&\red{73.83}
  \\
    \bottomrule
    \end{tabular}
    }

    \label{table3}
\end{table*}

\begin{table*}[!t]
\small
\centering
    \caption{Comparisons with SOTA methods for single-source multi-target domain generalization with the MLRSNet dataset as the source domain and remaining datasets as the target domains. Our method achieves better performance than the compared methods. We use \red{red} and \blu{blue} to highlight the first and second best scores.}
    \vspace{-2mm}
	\scalebox{0.95}{
\begin{tabular}{l c cccccccc}
    \toprule
    & \textbf{\ \ Source\ \ } & \multicolumn{8}{c}{\textbf{Target}} \\
    \cmidrule{2-2} \cmidrule(l){3-10}
   & MLRSNet & PatternNetv2 & RSSCN7v2 & AIDv2 & RSICDv2 & UCMv2 & WHURS19v2 & NWPUv2 & Average\\
    \midrule
    \ \  CoOp~\cite{zhou2022learning} & 72.53	&66.97	&69.07	&\red{67.13}	&\red{64.27}	&77.40	&\blu{85.20}	&71.17	&71.72\\
    \ \ CoCoOp~\cite{zhou2022conditional} \  & 71.70	&65.57	&69.37	&\red{67.13}	&\blu{62.73}	&75.70	&84.83	&70.97	&71.00\\
    \ \ MaPLe~\cite{khattak2023maple} & \blu{76.83}	&\red{68.03}	&\red{72.50}	&64.90	&59.73	&\red{78.93}	&83.07	&\blu{73.17}	&\blu{72.15}  \\
    \midrule
    \rowcolor{tabhighlight} \ \ Ours & \red{78.73}	&\blu{67.63}	&\blu{71.33}	&\blu{66.87}	&62.33	&\blu{78.33}	&\red{89.90}	&\red{73.67}	&\red{73.60}
  \\
    \bottomrule
    \end{tabular}
    }

    \label{table4}
\end{table*}

\subsection{Cross-Dataset Evaluation}
To evaluate the generalization capabilities of our proposed domain prompt learning method across datasets, we implement our approach on the MLRSNet and test the trained model directly on the remaining datasets. The experimental results and comparisons with other advanced prompt learning methods are presented in Table~\ref{table3}. Our proposed method achieves superior performance on the MLRSNet with a 1.90$\%$ performance improvement over the state-of-the-art. Notably, our method achieves the highest performance, reaching a 91.07$\%$ accuracy, on the WHURS19 dataset, suggesting a significant overlap in domain-specific information between MLRSNet and WHURS19. While our method does not attain top performance in every dataset, it ranks first in terms of average performance with a 1.41$\%$ accuracy improvement. These results affirm the enhanced cross-dataset generalization ability of our method.

\subsection{Domain Generalization}
To further validate the generalization capabilities of our proposed method, we conduct evaluations under the domain generalization setting like previous studies. The experimental results and comparisons with other advanced algorithms are detailed in Table~\ref{table4}. While the proposed method does not achieve top performance on each dataset, it excels with an average accuracy of 73.60$\%$, corresponding to a 1.35$\%$ improvement over MaPLe. Notably, the best performance is attained on the WHURS19V2 dataset, where our method exhibits a remarkable 4.70$\%$ performance boost. These results demonstrate that our proposed domain prompt learning effectively enhances the generalization and robustness of vision-language models, offering promising capabilities for domain generalization across datasets.

\begin{table}[t]
	\centering
	\scriptsize
	\footnotesize
	\renewcommand{\tabcolsep}{5.0mm}
	\caption{ The utilization of quaternion network (QN). Our method performs well over the compared methods due to successful cross-modal mining with QN.}
        \vspace{-2mm}
	\scalebox{1.0}{
		\begin{tabular}{c|ccc}
			\hline
			\multicolumn{1}{c|}{Methods}  &\multicolumn{1}{c}{Base}  &\multicolumn{1}{c}{Novel}  &\multicolumn{1}{c}{HM} \\
			\hline
      	\multicolumn{1}{c|}{Baseline} &93.93	&56.27	&70.38 \\
   		\multicolumn{1}{c|}{Ours w/o QN} &94.17	&63.53	&75.87\\
   		\multicolumn{1}{c|}{Ours}   &\textbf{95.67}	&\textbf{64.83}	&\textbf{77.29}\\
			\hline	
	\end{tabular} }
	\label{table5}
\end{table}

\begin{table}[t]
	\centering
	\scriptsize
	\footnotesize
	\renewcommand{\tabcolsep}{5.0mm}
	\caption{ The prompting of vision and language branches. PL: prompting language branch; PV: prompting vision branch. Our method performs better due to simultaneous PL and PV.}
        \vspace{-2mm}
	\scalebox{1.0}{
		\begin{tabular}{c|ccc}
			\hline
			\multicolumn{1}{c|}{Methods}  &\multicolumn{1}{c}{Base}  &\multicolumn{1}{c}{Novel}  &\multicolumn{1}{c}{HM} \\
			\hline
      	\multicolumn{1}{c|}{Baseline} &93.93	&56.27	&70.38 \\
   		\multicolumn{1}{c|}{PL} &95.23	&63.73	&76.36\\
            \multicolumn{1}{c|}{PV} &95.60	&64.23	&76.84\\
   		\multicolumn{1}{c|}{Ours (PL+PV)}   &\textbf{95.67}	&\textbf{64.83}	&\textbf{77.29}\\
			\hline	
	\end{tabular} }
	\label{table6}
\end{table}

\subsection{Ablation Studies}
\textbf{Quaternion Network.}
In order to examine the impact of different components of the proposed domain prompt learning, we carry out a series of ablation experiments. Initially, we study the effect of the quaternion neural network and its impact. Table~\ref{table5} shows that even without the quaternion network, our method still significantly improves the performance. This suggests that incorporating the domain-specific knowledge to transfer the VLMs from generalized to specialized is an efficient approach, and the quaternion network efficiently models the orthogonal cross-modal relationships, leading to further improved performance.

\vspace{1mm}\noindent \textbf{Prompting Vision and Language Branches.}
Table~\ref{table6} presents the experimental results of studying the impact of prompting vision and language branches. The results show that prompting either the language or vision branch alone leads to better performance due to improved domain-specific vision-language contrastive learning. By prompting both the vision and language branches complementarily, our proposed method successfully propagates domain-specific information into contrastive learning and significantly enhances recognition performance.

\vspace{1mm}\noindent \textbf{Prompting Depth.}
In domain prompt learning, the depth of prompting plays a crucial role, and we conduct experiments to study its impact. The results in Table~\ref{table7} show that performance improves as the depth increases until it reaches its highest point at a depth of 9. However, beyond that point, a further increase in depth leads to a decline in performance due to redundancy. Therefore, a depth of 9 is the appropriate choice for optimal performance.

\vspace{1mm}\noindent \textbf{Adding Noise.}
In order to mitigate potential overfitting problems, we devise an approach that introduces noise into the language branch. We assess the impact of this technique, and the results are presented in Table~\ref{table8}. Interestingly, our method still shows some performance improvements even in the absence of added noise. However, we have observed a decrease in accuracy when we add noise in the vision branch. We attribute this decline to the possibility of noise disrupting well-matched vision-language relationships. Our approach achieves the best performance, indicating that adding noise in the language branch is a beneficial strategy for addressing overfitting.

\begin{table}[t]
	\centering
	\scriptsize
	\footnotesize
	\renewcommand{\tabcolsep}{5.0mm}
	\caption{ Ablation studies of prompting depth. Our method performs better due to appropriate prompting depth.}
        \vspace{-2mm}
	\scalebox{1.0}{
		\begin{tabular}{c|ccc}
			\hline
			\multicolumn{1}{c|}{Depth}  &\multicolumn{1}{c}{Base}  &\multicolumn{1}{c}{Novel}  &\multicolumn{1}{c}{HM} \\
			\hline
      	\multicolumn{1}{c|}{3} &94.73	&60.7	&73.99\\
   		\multicolumn{1}{c|}{6} &95.63	&61.83	&75.10\\
            \multicolumn{1}{c|}{9} &\textbf{95.67}	&\textbf{64.83}	&\textbf{77.29}\\
   		\multicolumn{1}{c|}{12}  &95.57 &64.67	&77.14\\
			\hline	
	\end{tabular} }
	\label{table7}
\end{table}

\begin{table}[t]
	\centering
	\scriptsize
	\footnotesize
	\renewcommand{\tabcolsep}{5.0mm}
	\caption{ Ablation studies on the addition of noise. VB denotes the vision branch. LB denotes the language branch. Our method performs better due to correctly adding noise.}
        \vspace{-2mm}
	\scalebox{1.0}{
		\begin{tabular}{c|ccc}
			\hline
			\multicolumn{1}{c|}{Depth}  &\multicolumn{1}{c}{Base}  &\multicolumn{1}{c}{Novel}  &\multicolumn{1}{c}{HM} \\
			\hline
      	\multicolumn{1}{c|}{Baseline} &93.93	&56.27	&70.38 \\
   		\multicolumn{1}{c|}{w/o Noise} &95.43	&63.03	&75.92\\
            \multicolumn{1}{c|}{VB + Noise} &95.38	&62.95	&75.84\\
   		\multicolumn{1}{c|}{Ours (LB+ Noise)}  &\textbf{95.67}	&\textbf{64.83}	&\textbf{77.29}\\
			\hline	
	\end{tabular} }
	\label{table8}
\end{table}
\section{Conclusion}


While most advanced learning algorithms are designed to work with natural images, they often struggle to adapt to specific domains. To address this challenge, we harness domain-specific knowledge from the domain-specific foundation models and utilize quaternion networks to transfer the powerful recognition capabilities of VLMs from generalized to specialized domains. By analyzing cross-modal relationships between domain-specific vision features and contextual embeddings, the quaternion network seamlessly integrates the generalized contextual embeddings into the specialized space. Additionally, our approach incorporates domain-specific knowledge into the vision branch, enabling effective vision-language contrastive learning. We conduct extensive experiments on domain-specific datasets, and compelling experimental results demonstrate the superiority of our method.

{
    \small
    \bibliographystyle{ieeenat_fullname}
    \bibliography{main}
}


\end{document}